\documentclass[a4paper,fleqn]{cas-sc}

\usepackage{natbib}
\setcitestyle{authoryear}

\usepackage{times}
\usepackage{lipsum}
\usepackage{soul}
\usepackage{url}
\usepackage{float}
\usepackage[utf8]{inputenc}
\usepackage{import}
\usepackage{amsfonts}
\newcommand{\argmax}{\operatorname*{argmax}}

\usepackage{amsmath}
\usepackage{amsthm}
\usepackage{stmaryrd}
\usepackage{booktabs}
\usepackage{graphicx}
\usepackage{multirow}
\usepackage{breakcites}
\usepackage{multirow}
\usepackage{boldline}
\usepackage{wrapfig}

\setcitestyle{square}
\let\cite\citep

\usepackage{xcolor,pifont}
\newcommand*\colourcheck[1]{%
  \expandafter\newcommand\csname #1check\endcsname{\textcolor{#1}{\ding{52}}}%
}
\colourcheck{blue}
\colourcheck{green}
\colourcheck{red}

\usepackage{tikz}
\usetikzlibrary{arrows,shapes,positioning,shadows,trees}

\tikzset{
  basic/.style  = {draw, text width=2cm, drop shadow, font=\sffamily, rectangle, text = black},
  root/.style   = {basic, rounded corners=2pt, thin, align=center,
                   fill=gray!60},
  level 2/.style = {basic, rounded corners=6pt, thin,align=center, fill=gray!30,
                   text width=8em},
  level 3/.style = {basic, thin, align=left, fill=gray!10, text width=5.9em},
  level 4/.style = {basic,  rounded corners=2pt, thin, align=center, fill=gray!60,text width=8em},
  level 5/.style = {basic, thin, align=left, fill=gray!20, text width=7em},
}

\theoremstyle{definition}
\newtheorem{definition}{Definition}

\begin{document}
\let\WriteBookmarks\relax
\def\floatpagepagefraction{1}
\def\textpagefraction{.001}
\shorttitle{Reinforcement Learning for Combinatorial Optimization}
\shortauthors{Mazyavkina et al.}

\title[mode = title]{Reinforcement Learning for Combinatorial Optimization: A Survey} 

\author[skol]{Nina Mazyavkina}[orcid=0000-0002-0874-4165]\corref{mycorrespondingauthor}
\cortext[mycorrespondingauthor]{Corresponding author}
\ead{nina.mazyavkina@skoltech.ru}

\author[zyfra]{Sergey Sviridov}
\author[criteo,skol]{Sergei Ivanov}
\author[skol]{Evgeny Burnaev}

\address[skol]{Skolkovo Institute of Science and Technology, Russia}
\address[zyfra]{Zyfra, Russia}
\address[criteo]{Criteo AI Lab, France}

\begin{abstract}
Many traditional algorithms for solving combinatorial optimization problems involve using hand-crafted heuristics that sequentially construct a solution. Such heuristics are designed by domain experts and may often be suboptimal due to the hard nature of the problems. Reinforcement learning (RL) proposes a good alternative to automate the search of these heuristics by training an agent in a supervised or self-supervised manner. In this survey, we explore the recent advancements of applying RL frameworks to hard combinatorial problems. Our survey provides the necessary background for operations research and machine learning communities and showcases the works that are moving the field forward. We juxtapose recently proposed RL methods, laying out the timeline of the improvements for each problem, as well as we make a comparison with traditional algorithms, indicating that RL models can become a promising direction for solving combinatorial problems. 
\end{abstract}

\begin{keywords}
reinforcement learning, operations research, combinatorial optimization, value-based methods, policy-based methods
\end{keywords}

\maketitle

\section{Introduction}
\label{sec:intro}
Optimization problems are concerned with finding optimal configuration or "value" among different possibilities, and they naturally fall into one of the two buckets: configurations with continuous and with discrete variables. For example, finding a solution to a convex programming problem is a continuous optimization problem, while finding the shortest path among all paths in a graph is a discrete optimization problem. Sometimes the line between the two can not be drawn that easily. For example, the linear programming task in the continuous
space can be regarded as a discrete combinatorial problem because its solution lies in a finite set of vertices of the convex polytope as it has been demonstrated by Dantzig’s algorithm \cite{dantzig2006linear}. Conventionally, optimization problems in the discrete space are called combinatorial optimization (CO) problems and, typically, have different types of solutions comparing to the ones in the continuous space. One can formulate a CO problem as follows: 

\begin{definition}
\label{def:co}
    Let $V$ be a set of elements and $f: V \mapsto \mathbb{R}$ be a cost function. \textit{Combinatorial optimization problem} aims to find an optimal value of the function $f$ and any corresponding optimal element that achieves that optimal value on the domain $V$.
\end{definition}

Typically the set $V$ is finite, in which case there is a global optimum, and, hence, a trivial solution exists for any CO problem by comparing values of all elements $v \in V$. Note that the definition \ref{def:co} also includes the case of decision problems, when the solution is binary (or, more generally, multi-class), by associating a higher cost for the wrong answer than for the right one. One common example of a combinatorial problem is Travelling Salesman Problem (TSP). The goal is to provide the shortest route that visits each vertex and returns to the initial endpoint, or, in other words, to find a Hamiltonian circuit $H$ with minimal length in a fully-connected weighted graph. In this case, a set of elements is defined by all Hamiltonian circuits, i.e. $V = \{\text{all Hamiltonian paths}\}$, and the cost associated with each Hamiltonian circuit is the sum of the weights $w(e)$ of the edges $e$ on the circuit, i.e. $f(H)=\sum\limits_{e\in H} w(e)$. Another example of CO problem is Mixed-Integer Linear Program (MILP), for which the objective is to minimize $c^{\top}x$ for a given vector $c\in \mathbb{R}^d$ such that the vector $x \in \mathbb{Z}^d$ satisfies the constraints $Ax \le b$ for the parameters $A$ and $b$.

Many CO problems are NP-hard and do not have an efficient polynomial-time solution. As a result, many algorithms that solve these problems either approximately or heuristically have been designed. One of the emerging trends of the last years is to solve CO problems by training a machine learning (ML) algorithm. For example, we can train ML algorithm on a dataset of already solved TSP instances to decide on which node to move next for new TSP instances. A particular branch of ML that we consider in this survey is called reinforcement learning (RL) that for a given CO problem defines an environment and the agent that acts in the environment constructing a solution. 

In order to apply RL to CO, the problem is modeled as a sequential decision-making process, where the agent interacts with the environment by performing a sequence of actions in order to find a solution. \textit{Markov Decision Process (MDP)} provides a widely used mathematical framework for modeling this type of problems \cite{Bel}.
\begin{definition}
MDP can be defined as a tuple $M = \langle S, A, R, T, \gamma, H \rangle$, where
\begin{itemize}
\item $S$ - \emph{state space} $\bold{s}_t \in S$. State space for combinatorial optimization problems in this survey is typically defined in one of two ways.  One group of approaches constructs solutions incrementally define it as a set of partial solutions to the problem (e.g. a partially constructed path for TSP problem). The other group of methods starts with a suboptimal solution to a problem and iteratively improves it (e.g. a suboptimal tour for TSP). 
\item $A$ - \emph{action space} $\bold{a}_t \in A$. Actions represent addition to partial or changing complete solution (e.g. changing the order of nodes in a tour for TSP);
\item $R$ - \emph{reward function} is a mapping from states and actions into real numbers $R: S\times A \xrightarrow{} \mathbb{R}$. Rewards indicate how action chosen in particular state improves or worsens a solution to the problem (i.e. a tour length for TSP);
\item $T$ - \emph{transition function} $T(\bold{s}_{t+1} | \bold{s}_t, \bold{a}_t)$ that governs transition dynamics from one state to another in response to action. In combinatorial optimization setting transition dynamics is usually deterministic and known in advance;
\item $\gamma$ - scalar \emph{discount factor}, $0 < \gamma \leq 1$. Discount factor encourages the agent to account more for short-term rewards;
\item $H$ - \emph{horizon}, that defines the length of the episode, where episode is defined as a sequence $\{s_t, a_t, s_{t+1}, a_{t+1}, s_{t+2}, ...\}_{t=0}^{H}$.  For methods that construct solutions incrementally episode length is defined naturally by number of actions performed until solution is found. For iterative methods some artificial stopping criteria are introduced.
\end{itemize}
\end{definition}

The goal of an agent acting in Markov Decision Process is to find \textit{a policy function} $\pi(s)$ that maps states into actions. Solving MDP means finding the \textit{optimal policy} that maximizes the expected cumulative discounted sum of rewards:

\begin{equation}\label{opt_pi}
	    \pi^* = \argmax_{\pi} \mathbb{E}[\sum\limits_{t=0}^{H} \gamma^t R(s_t,a_t)],
\end{equation}

\begin{figure}[h!]
\vskip -0.1in
\begin{center}
\centerline{\includegraphics[scale=0.65]{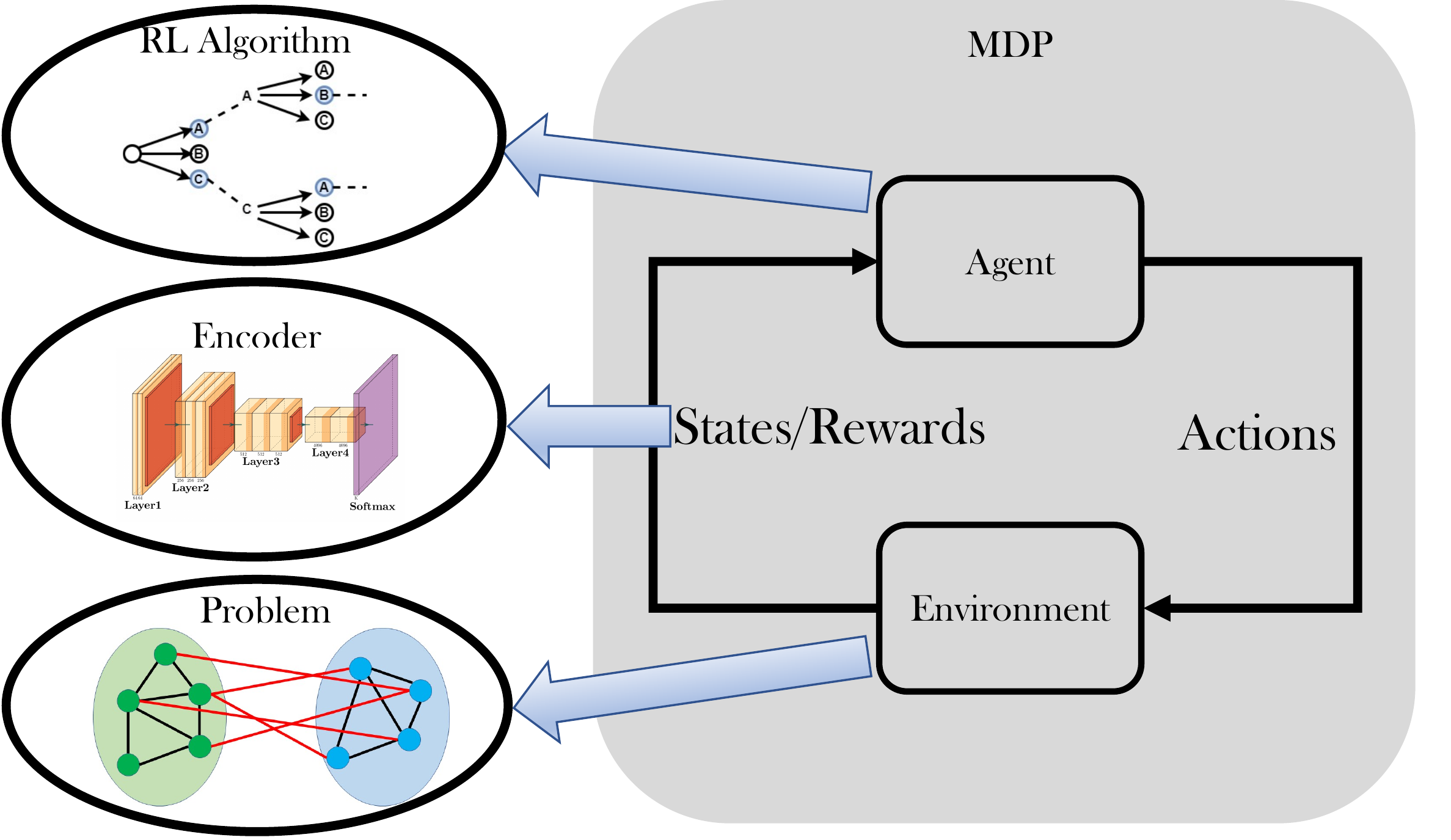}}
\caption{Solving a CO problem with the RL approach requires formulating MDP. The environment is defined by a particular instance of CO problem (e.g. Max-Cut problem). States are encoded with a neural network model (e.g. every node has a vector representation encoded by a graph neural network). The agent is driven by an RL algorithm (e.g. Monte-Carlo Tree Search) and makes decisions that move the environment to the next state (e.g. removing a vertex from a solution set).}
\label{fig:pipeline}
\end{center}
\vskip -0.2in
\end{figure}

Once MDP has been defined for a CO problem we need to decide how the agent would search for the optimal policy $\pi^*$. Broadly, there are two types of RL algorithms:
\begin{itemize}
\item \textit{Value-based methods} first compute the value action function $Q^{\pi}(s, a)$ as the expected reward of a policy $\pi$ given a state $s$ and taking an action $a$. Then the agent’s policy corresponds to picking an action that maximizes $Q^{\pi}(s, a)$ for a given state. The main difference between value-based approaches is in how to estimate $Q^{\pi}(s, a)$ accurately and efficiently.
\item \textit{Policy-based methods} directly model the agent’s policy as a parametric function $\pi_{\theta}(s)$. By collecting previous decisions that the agent made in the environment, also known as experience, we can optimize the parameters $\theta$ by maximizing the final reward \ref{opt_pi}. The main difference between policy-based methods is in optimization approaches for finding the function $\pi_{\theta}(s)$ that maximizes the expected sum of rewards.
\end{itemize}

As can be seen, RL algorithms depend on the functions that take as input the states of MDP and outputs the actions' values or actions. States represent some information about the problem such as the given graph or the current tour of TSP, while Q-values or actions are numbers. Therefore an RL algorithm has to include an \textit{encoder}, i.e., a function that encodes a state to a number. Many encoders were proposed for CO problems including recurrent neural networks, graph neural networks, attention-based networks, and multi-layer perceptrons.

To sum up, a pipeline for solving CO problem with RL is presented in Figure~\ref{fig:pipeline}. A CO problem is first reformulated in terms of MDP, i.e., we define the states, actions, and rewards for a given problem. We then define an encoder of the states, i.e. a parametric function that encodes the input states and outputs a numerical vector (Q-values or probabilities of each action). The next step is the actual RL algorithm that determines how the agent learns the parameters of the encoder and makes the decisions for a given MDP. After the agent has selected an action, the environment moves to a new state and the agent receives a reward for the action it has made. The process then repeats from a new state within the allocated time budget. Once the parameters of the model have been trained, the agent is capable of searching the solutions for unseen instances of the problem.

Our work is motivated by the recent success in the application of the techniques and methods of the RL field to solve CO problems. Although many practical combinatorial optimization problems can be, in principle, solved by reinforcement learning algorithms with relevant literature existing in the operations research community, we will focus on RL approaches for CO problems. This survey covers the most recent papers that show how reinforcement learning algorithms can be applied to reformulate and solve some of the canonical optimization problems, such as Travelling Salesman Problem (TSP), Maximum Cut (Max-Cut) problem, Maximum Independent Set (MIS), Minimum Vertex Cover (MVC), Bin Packing Problem (BPP).

\textbf{Related work.}  Some of the recent surveys also describe the intersection of machine learning and combinatorial optimization. This way a comprehensive survey by \cite{bengio2020machine} has summarized the approaches that solve CO problems from the perspective of the general ML, and the authors have discussed the possible ways of the combination of the ML heuristics with the existing off-the-shelf solvers. Moreover, the work by \cite{zhou2018graph}, which is devoted to the description and possible applications of GNNs, has described the progress on the CO problems' formulation from the GNN perspective in one of its sections. Finally, the more recent surveys by \cite{vesselinova2020learning} and \cite{guo2019solving}, describe the latest ML approaches to solving the CO tasks, in addition to possible applications of such methods. We note that our survey is complementary to the existing ones as we focus on RL approaches, provide necessary background and classification of the RL models, and make a comparison between different RL methods and existing solutions. 

\textbf{Paper organization.} The remainder of this survey is organized as follows. In section \ref{sec:background}, we provide a necessary background including the formulation of CO problems, different encoders, and RL algorithms that are used for solving CO with RL. In section \ref{sec:taxonomy} we provide a classification of the existing RL-CO methods based on the popular design choices such as the type of RL algorithm. In section \ref{sec:problems} we describe the recent RL approaches for the specific CO problems, providing the details about the formulated MDPs as well as their influence on other works. In section \ref{sec:comparison} we make a comparison between the RL-CO works and the existing traditional approaches. We conclude and provide future directions in section \ref{sec:conclusion}.

\section{Background}
\label{sec:background}
In this section, we provide definitions of combinatorial problems, state-of-the-art algorithms and heuristics that solve these problems. We also describe machine learning models that encode states of CO problems for an RL agent. Finally, we categorize popular RL algorithms that have been employed recently for solving CO problems.

\subsection{Combinatorial Optimization Problems}

We start by considering mixed-integer linear programs (MILP) -- a constrained optimization problem, to which many practical applications can be reduced. Several industrial optimizers (e.g. \citep{manual1987ibm, gleixner2017scip, gurobi, sagemath, GLPK, schrage1986linear}) exist that use a branch-and-bound technique to solve the MILP instance.

\begin{definition}[Mixed-Integer Linear Program (MILP) \citep{wolsey1998integer}]
A mixed-integer linear program is an optimization problem of the form
\[
\underset{\mathbf{x}}{\arg \min }\left\{\mathbf{c}^{\top} \mathbf{x} \text{ } | \text{ } \mathbf{A} \mathbf{x} \leq \mathbf{b}, \quad \mathbf{0} \leq \mathbf{x}, \quad \mathbf{x} \in \mathbb{Z}^{p} \times \mathbb{R}^{n-p}\right\},
\]
where $\mathbf{c} \in \mathbb{R}^{n}$ is the objective coefficient vector, $\mathbf{A} \in \mathbb{R}^{m \times n}$ is the constraint coefficient matrix, $\mathbf{b} \in \mathbb{R}^{m}$ is the constraint vector, and $p \leq n$ is the number of integer variables.
\end{definition}


Next, we provide formulations of the combinatorial optimization problems, their time complexity, and the state-of-the-art algorithms for solving them.


\begin{definition}[Traveling Salesman Problem (TSP)]
\label{def:tsp}
Given a complete weighted graph $G=(V,E)$, find a tour of minimum total weight, i.e. a cycle of minimum length that visits each node of the graph exactly once. 
\end{definition}

TSP is a canonical example of a combinatorial optimization problem, which has found applications in planning, data clustering, genome sequencing, etc. \citep{applegate2006traveling}. TSP problem is NP-hard \citep{papadimitriou1998combinatorial}, and many exact, heuristic, and approximation algorithms have been developed, in order to solve it. The best known exact algorithm is the Held–Karp algorithm \citep{held1962dynamic}. Published in 1962, it solves the problem in time ${\displaystyle O(n^{2}2^{n})}$, which has not been improved in the general setting since then. TSP can be formulated as a MILP instance \citep{dantzig1954solution, miller1960integer}, which allows one to apply MILP solvers, such as Gurobi \citep{gurobi}, in order to find the exact or approximate solutions to TSP. Among them, Concorde \citep{applegate2006traveling} is a specialized TSP solver that uses a combination of cutting-plane algorithms with a branch-and-bound approach. Similarly, an extension of the Lin-Kernighan-Helsgaun TSP solver (LKH3) \citep{helsgaun2017extension}, which improves the Lin-Kernighan algorithm \citep{lin1973effective}, is a tour improvement method that iteratively decides which edges to rewire to decrease the tour length. More generic solvers that avoid local optima exist such as OR-Tools \citep{ortools} that tackle vehicle routing problems through local search algorithms and metaheuristics. In addition to solvers, many heuristic algorithms have been developed, such as Christofides-Serdyukov algorithm \citep{christofides1976worst, van2020historical},  the Lin-Kernighan-Helsgaun heuristic \citep{helsgaun2000effective}, 2-OPT local search \citep{mersmann2012local}. \citep{applegate2006traveling} provides an extensive overview of various approaches to TSP.

\begin{definition}[Maximum Cut Problem (Max-Cut)]
\label{def:maxcut}
    Given a graph $G = (V,E)$, find a subset of vertices $S \subset V$ that maximizes a cut $C(S, G) = \sum_{i\in S,j \in V\setminus S} w_{ij}$ where $w_{ij} \in W$ is the weight of the edge-connecting vertices $i$ and $j$.
\end{definition}

Max-Cut solutions have found numerous applications in real-life problems including protein folding \citep{perdomo2012finding}, financial portfolio management \citep{elsokkary2017financial}, and finding the ground state of the Ising Hamiltonian in physics \citep{barahona1982computational}. Max-Cut is an NP-complete problem \citep{karp1972reducibility}, and, hence, does not have a known polynomial-time algorithm. Approximation algorithms exist for Max-Cut, including deterministic 0.5-approximation \citep{Mitzenmacher, gonzalez2007handbook} and randomized 0.878-approximation \citep{goemans1995improved}. Industrial solvers can be used to find a solution by applying the branch-and-bound routines. In particular, Max-Cut problem can be transformed into a quadratic unconstrained binary optimization problem and solved by CPLEX \citep{manual1987ibm}, which takes within an hour for graph instances with hundreds of vertices \citep{barrett2019exploratory}. For larger instances several heuristics using the simulated annealing technique have been proposed that could scale to graphs with thousands of vertices \citep{yamamoto2017coherent,tiunov2019annealing,leleu2019destabilization}. 

\begin{definition}[Bin Packing Problem (BPP)]
\label{def:3dbpp}
Given a set ${\displaystyle I}$ of items, a size ${\displaystyle s(i)\in \mathbb {Z} ^{+}}$ for each ${\displaystyle i\in I}$, and a positive integer bin capacity ${\displaystyle B}$, find a partition of $I$ into disjoint sets ${\displaystyle I_{1},\dots ,I_{K}}$ such that the sum of the sizes of the items in each ${\displaystyle I_{j}}$ is less or equal than ${\displaystyle B}$ and $K$ has the smallest possible value.
\end{definition}

There are other variants of BPP such as 2D, 3D packing, packing with various surface area, packing by weights, and others \citep{wu2010three}. This CO problem has found its applications in many domains such as resource optimization, logistics, and circuit design \citep{kellerer2004multidimensional}. BPP is an NP-complete problem with many approximation algorithms proposed in the literature. First-fit decreasing (FFD) and best-fit decreasing (BFD) are two simple approximation algorithms that first sort the items in the decreasing order of their costs and then assign each item to the first (for FFD) or the fullest (for BFD) bin that it fits into. Both FFD and BFD run in $O(n\log n)$ time and have $11/9$ asymptotic performance guarantee \citep{korte2012combinatorial}. Among exact approaches, one of the first attempts has been the Martello-Toth algorithm that works under the branch-and-bound paradigm \citep{martello1990bin, martello1990lower}. In addition, several recent improvements have been proposed \citep{schreiber2013improved, korf2003improved} which can run on instances with hundreds of items. Alternatively, BPP can be formulated as a MILP instance \citep{wu2010three, chen1995analytical} and solved using standard MILP solvers such as Gurobi \citep{gurobi} or CPLEX \citep{manual1987ibm}.

\begin{definition}[Minimum Vertex Cover (MVC)]
\label{def:minvertex}
Given a graph $G = (V,E)$, find a subset of nodes $ S \subset V$, such that every edge is covered, i.e. $(u, v) \in E \iff u \in S \text{ or } v \in S$, and $|S|$ is minimized.
\end{definition}

Vertex cover optimization is a fundamental problem with applications to computational biochemistry \citep{lancia2001snps} and computer network security \citep{filiol2007combinatorial}. There is a na\"ive approximation algorithm with a factor 2, which works by adding both endpoints of an arbitrary edge to the solution and then removing this endpoints from the graph \citep{papadimitriou1998combinatorial}. A better approximation algorithm with a factor of $2-\Theta \left(1/{\sqrt {\log |V|}}\right)$ is known \citep{karakostas2009better}, although, it has been shown that MVC cannot be approximated within a factor $\sqrt {2}-\varepsilon$ for any $\varepsilon > 0$ \citep{dinur2005hardness, subhash2018pseudorandom}. The problem can formulated as an integer linear program (ILP) by minimizing $\sum _{v\in V}c(v)x_{v}$, where $x_{v}\in \{0,1\}$ denotes whether a node $v$ with a weight $c(v)$ is in a solution set, subject to $x_{u}+x_{v}\geq 1$. Solvers such as CPLEX \citep{manual1987ibm} or Gurobi \citep{gurobi} can be used to solve the ILP formulations with hundreds of thousands of nodes \citep{akiba2016branch}. 

\begin{definition}[Maximum Independent Set (MIS)]
\label{def:mis}
Given a graph $G(V, E)$ find a subset of vertices $S \subset V$, such that no two vertices in $S$ are connected by an edge of $E$, and $|S|$ is minimized.
\end{definition}

MIS is a popular CO problem with applications in classification theory, molecular docking, recommendations, and more \citep{feo1994greedy, gardiner2000graph, agrawal1996fast}. As such the approaches of finding the solutions for this problem have received a lot of attention from the academic community. It is easy to see that the complement of an independent set in a graph $G$ is a vertex cover in $G$ and a clique in a complement graph $\bar{G}$, hence, the solutions to a minimum vertex cover in $G$ or a maximum clique in $\overline{G}$ can be applied to solve the MIS problem. The running time of the brute-force algorithm is $O(n^2 2^n)$, which has been improved by \citep{tarjan1977finding} to $O(2^{n/3})$, and recently to the best known bound $O(1.1996^n)$ with polynomial space \citep{xiao2017exact}. To cope with medium and large instances of MIS several local search and evolutionary algorithms have been proposed. The local search algorithms maintain a solution set, which is iteratively updated by adding and removing nodes that improve the current objective value \citep{andrade2008fast, katayama2005effective, hansen2004variable,pullan2006dynamic}. In contrast, the evolutionary algorithms maintain several independent sets at the current iterations which are then merged or pruned based on some fitness criteria \citep{lamm2015graph, borisovsky2003experimental, back1994evolutionary}. Hybrid approaches exist that combine the evolutionary algorithms with the local search, capable to solve instances with hundreds of thousands of vertices \citep{lamm2016finding}.

In order to approach the outlined problems with reinforcement learning, we must represent the graphs, involved in the problems, as vectors that can be further provided as an input to a machine learning algorithm. Next, we discuss different approaches for learning the representations of these problems.

\subsection{Encoders}
\label{subsec:encoders}
In order to process the input structure $S$ (e.g. graphs) of CO problems, we must present a mapping from $S$ to a $d$-dimensional space $\mathbb{R}^d$. We call such a mapping an \textit{encoder} as it encodes the original input space. The encoders vary depending on the particular type of the space $S$ but there are some common architectures that researchers have developed over the last years to solve CO problems. 

\begin{wrapfigure}{r}{0.5\textwidth}
\vskip -0.1in
  \begin{center}
    \includegraphics[width=0.35\textwidth]{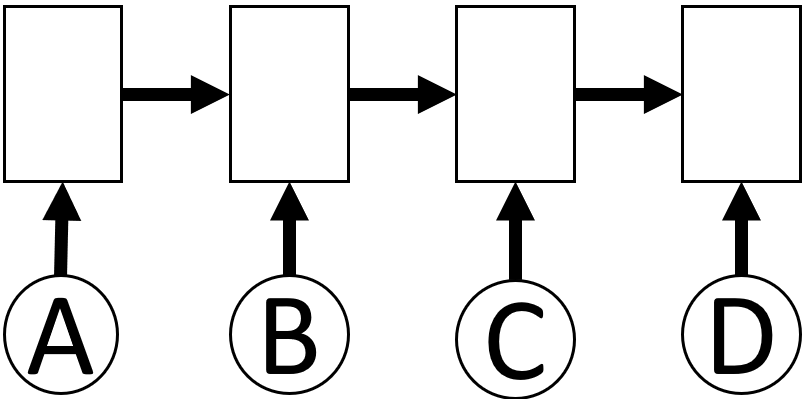}
  \end{center}
  \vskip -0.2in
  \caption{The scheme for a recurrent neural network (RNN). Each box represents an encoding function. Each element in the sequence is encoded using its initial representation and the output of the model at the previous step. RNN parameters are shared across all elements of the sequence.}
  \label{fig:rnn}
\end{wrapfigure}

The first frequently used architecture is a \textit{recurrent neural network} (RNN). RNNs can operate on sequential data, encoding each element of the sequence into a vector. In particular, the RNN is composed of the block of parameters that takes as an input the current element of the sequence and the previous output of the RNN block and outputs a vector that is passed to the next element of the sequence. For example, in the case of TSP, one can encode a tour of TSP by applying RNN to the current node (e.g. initially represented by a constant $d$-dimensional vector) and the output of the RNN on the previous node of the tour. One can stack multiple blocks of RNNs together making the neural network deep. Popular choices of RNN blocks are a Long Short-Term Memory (LSTM) unit \citep{hochreiter1997long} and Gated Recurrent Unit (GRU) \citep{cho2014learning}, which tackle the vanishing gradient problem \citep{goodfellow2016deep}. 

One of the fundamental limitations of RNN models is related to the modeling of the long-range dependencies: as the model takes the output of the last time-step it may “forget” the information from the previous elements of the sequence. \textit{Attention models} fix this by forming a connection not just to the last input element, but to all input elements. Hence, the output of the attention model depends on the current element of the sequence and all previous elements of the sequence. In particular, similarity scores (e.g. dot product) are computed between the input element and each of the previous elements, and these scores are used to determine the weights of the importance of each of the previous elements to the current element. Attention models has recently gained the superior performance on language modeling tasks (e.g. language translation) \citep{vaswani2017attention} and have been applied to solving CO problems (e.g. for building incrementally a tour for TSP).

\begin{wrapfigure}{r}{0.5\textwidth}
\vskip -0.1in
  \begin{center}
    \includegraphics[width=0.35\textwidth]{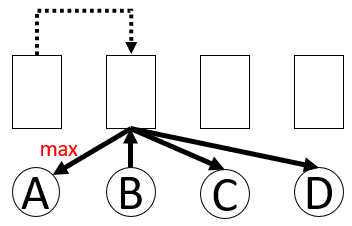}
  \end{center}
  \vskip -0.2in
  \caption{The scheme for a pointer network. Element "B" in the sequence first computes similarity scores to all other elements. Next we encode the representation of "B" using the element with maximum value ("A" in this case, dashed). This process is then repeated for other elements in the sequence.}
  \label{fig:pointer}
\end{wrapfigure}

Note that the attention model relies on modeling dependencies between each pair of elements in the input structure, which can be inefficient if there are only few relevant dependencies. One simple extension of the attention model is a \textit{pointer network} (PN) \citep{vinyals2015pointer}. Instead of using the weights among all pairs for computation of the influence of each input element, the pointer networks use the weights to select a single input element that will be used for encoding. For example, in Figure~\ref{fig:pointer} the element "A" has the highest similarity to the element "B", and, therefore, it is used for computation of the representation of element "B" (unlike attention model, in which case the elements "C" and "D" are also used). 

Although these models are general enough to be applied to various spaces $S$ (e.g. points for TSP), many CO problems studied in this paper are associated with the graphs. A natural continuation of the attention models to the graph domain is a \textit{graph neural network} (GNN). Initially, the nodes are represented by some vectors (e.g. constant unit vectors). Then, each node’s representation is updated depending on the local neighborhood structure of this node. In the most common message-passing paradigms, adjacent nodes exchange their current representations in order to update them in the next iteration. One can see this framework as a generalization of the attention model, where the elements do not attend to all of the other elements (forming a fully-connected graph), but only to elements that are linked in the graph. Popular choices of GNN models include a Graph Convolutional Network (GCN) \citep{kipf2017semi}, a Graph Attention Network (GAT) \citep{velivckovic2017graph}, a a Graph Isomorphism Network (GIN) \citep{xu2018powerful}, Structure-to-Vector Network (S2V) \citep{dai2016discriminative}. 

While there are many intrinsic details about all of these models, at a high level it is important to understand that all of them are the differentiable functions optimized by the gradient descent that return the encoded vector representations, which next can be used by the RL agent.

\subsection{Reinforcement Learning Algorithms}
\label{sec:rl-algo}
In the introduction section~\ref{sec:intro} we gave the definitions of an MDP, which include the states, actions, rewards, and transition functions. We also explained what the policy of an agent is and what is the optimal policy. Here we will deep-dive into the RL algorithms that search for the optimal policy of an MDP. 

\begin{figure}
\centering
\includegraphics[width=\textwidth]{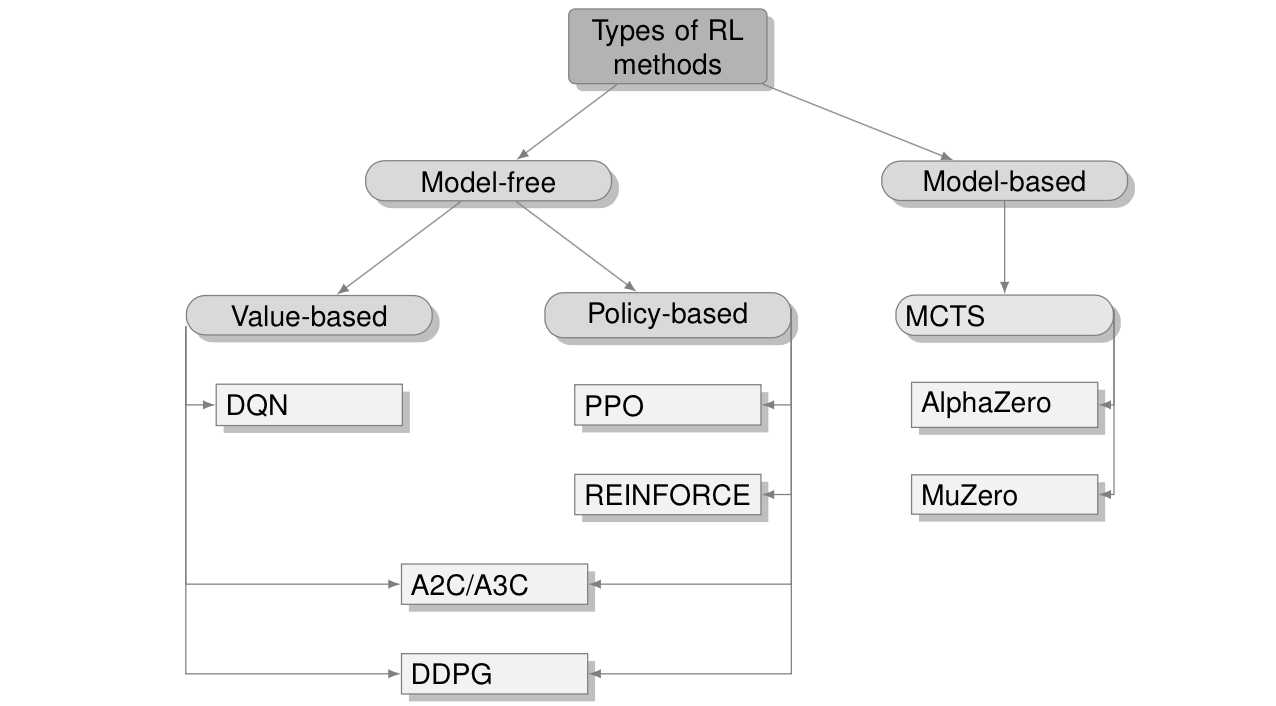}
\caption{A classification of reinforcement learning methods.}
\label{fig:rl_methods}
\end{figure}

Broadly, the RL algorithms can be split into the model-based and model-free categories (Figure~\ref{fig:rl_methods}). 

\begin{itemize}
    \item \textit{Model-based methods} focus on the environments, where transition functions are known or can be learned, and can be utilized by the algorithm when making decisions. This group includes \textit{Monte-Carlo Tree Search} (MCTS) algorithms such as AlphaZero \citep{silver2016mastering} and MuZero \citep{schrittwieser2019mastering}. 
    
    \item \textit{Model-free methods} do not rely on the availability of the transition functions of the environment and utilize solely the experience collected by the agent. 
\end{itemize}

Furthermore, model-free methods can be split into two big families of RL algorithms -- policy-based and value-based methods. This partition is motivated by the way of deriving a solution of an MDP. In the case of policy-based methods, a policy is approximated directly, while value-based methods focus on approximating a value function, which is a measure of the quality of the policy for some state-action pair in the given environment. Additionally, there are RL algorithms that combine policy-based methods with value-based methods. The type of methods that utilize such training procedure is called actor-critic methods \cite{sutton2000policy,mnih2016asynchronous}. The basic principle behind these algorithms is for the critic model to approximate the value function, and for the actor model to approximate policy. Usually to do this, both actor and critic, use the policy and value-based RL, mentioned above. This way, the critic provides the measure of how good the action taken by the actor has been, which allows to appropriately adjust the learnable parameters for the next train step.

Next, we formally describe the value-based, policy-based, and MCTS approaches and the corresponding RL algorithms that have been used to solve CO problems.

\subsubsection{Value-based methods}
\label{sec:value}
As it has been mentioned earlier, the main goal of all reinforcement learning methods is to find a policy, which would consistently allow the agent to gain a lot of rewards. Value-based reinforcement learning methods focus on finding such policy through the approximation of a value function $V(s)$ and an action-value function $Q(s,a)$. In this section, we will define both of these functions, which value and action-value functions can be called optimal, and how can we derive the optimal policy, knowing the optimal value functions.

\begin{definition} \textit{Value function} of a state $s$ is the expectation of the future discounted rewards, when starting from the state $s$ and following some policy $\pi$:
\begin{equation}
V^{\pi}(s) = \mathbb{E}\bigg[\sum_{t=0}^{\infty}\gamma^{t} r(s_t) | \pi, s_0 = s\bigg]
\end{equation}

\end{definition}

The notation $V^{\pi}$ here and in the following sections means that the value function $V$ is defined with respect to the policy $\pi$. It is also important to note, that the value of a terminal state in the case of a finite MDP equals $0$.

 At the same time, it can be more convenient to think of the value function as the function depending not only on the state but also on the action. 
 \begin{definition}
 \textit{Action-value function} $Q(s,a)$ is the expectation of the future discounted rewards, when starting from the state $s$, taking the action $a$ and then following some policy $\pi$:
 \begin{equation}
 Q^{\pi}(s,a) = \mathbb{E} \bigg[\sum_{t=0}^{\infty}\gamma^t r(s_t, a_t) |\pi, s_0 = s, a_0 = a\bigg].    
 \end{equation}
 \end{definition}
 It is also clear that $V^{\pi}(s)$ can be interpreted in terms of the $Q^{\pi}(s,a)$ as:
$$
V^{\pi}(s) = \max_{a}Q^{\pi}(s,a).
$$

From the definition of a value function comes a very important recursive property, representing the relationship between the value of the state $V^{\pi}(s)$ and the values of the possible following states $V^{\pi}(s')$, which lies at the foundation of many value-based RL methods. This property can be expressed as an equation, called the Bellman equation \cite{bellman1952theory}:
\begin{equation}
\label{eq:bellman-value}
  V^{\pi}(s) = r(s) + \gamma \sum_{s'}T(s,\pi(s),s')V^{\pi}(s').  
\end{equation}

The Bellman equation can be also rewritten in terms of the action-value function $Q^{\pi}(s,a)$ in the following way:
\begin{equation}
\label{eq:bellman-action}
Q^{\pi}(s,a) = r(s,a) + \gamma \sum_{s'}T(s,a,s')\max_{a'}Q^{\pi}(s',a').
\end{equation}

At the beginning of this section, we have stated that the goal of all of the RL tasks is to find a policy, which can accumulate a lot of rewards. This means that one policy can be better than (or equal to) the other if the expected return of this policy is greater than the one achieved by the other policy: $\pi' \geq \pi$. Moreover, by the definition of a value function, we can claim that $\pi' \geq \pi$ if and only if $V^{\pi'}(s) \geq V^{\pi}(s)$ in all states $s \in S$. 

Knowing this relationship between policies, we can state that there is a policy that is better or equal to all the other possible policies. This policy is called an \textit{optimal policy} $\pi^{*}$. Evidently, the optimality of the action-value and value functions is closely connected to the optimality of the policy they follow. This way, the value function of an MDP is called \textit{optimal} if it is the maximum of value functions across all policies:
$$
V^{*}(s) = \max_{\pi} V^{\pi}(s), \forall s \in S.
$$

Similarly, we can give the definition to the \textit{optimal action-value function} $Q^{*}(s,a)$:
$$
Q^{*}(s,a) = \max_{\pi} Q^{\pi}(s,a), \forall s \in S,  \forall a \in A.
$$

Given the Bellman equations (\ref{eq:bellman-value}) and (\ref{eq:bellman-action}), one can derive the optimal policy if the action-value or value functions are known. In the case of a value function $V^{*}(s)$, one can find optimal actions by doing the greedy one-step search: picking the actions that correspond to the maximum value $V^{*}(s)$ in the state $s$ computed by the Bellman equation (\ref{eq:bellman-value}). On the other hand, in the case of the action-value function one-step search is not needed. For each state $s$ we can easily find such action $a$ that maximizes the action-function, as in order to do that we just need to compute $Q^{*}(s,a)$. This way, we do not need to know any information about the rewards and values in the following states $s'$ in contrast with the value function.

Therefore, in the case of value-based methods, in order to find the optimal policy, we need to find the optimal value functions. Notably, it is possible to explicitly solve the Bellman equation, i.e. find the optimal value function, but only in the case when the transition function is known. In practice, it is rarely the case, so we need some methods to approximate the Bellman's equation solution. 

\begin{itemize}
    \item \textbf{Q-learning.} One of the popular representatives of the approximate value-based methods is Q-learning \citep{watkins1992q} and its deep variant Deep Q-learning \citep{mnih2015human}. In Q-learning, the action-value function $Q(s,a)$ is iteratively updated by learning from the collected experiences of the current policy. It has been shown in \cite[Theorem~3]{sutton1988learning} that the function updated by such a rule converges to the optimal value function.
    \item \textbf{DQN.}  With the rise of Deep Learning, neural networks (NNs) have proven to achieve state-of-the-art results on various datasets by learning useful function approximations through the high-dimensional inputs. This led researchers to explore the potential of NNs' approximations of the Q-functions. Deep Q-networks (DQN) \citep{mnih2015human} can learn the policies directly using end-to-end reinforcement learning. The network approximates Q-values for each action depending on the current input state. In order to stabilize the training process, authors have used the following formulation of the loss function:

\begin{equation}\label{loss_dqn}
  L(\theta_i) = \mathbb{E}_{(s,a,r,s') \sim D}\big[\big(r + \gamma \max_{a'}Q_{\theta_i^-}(s',a') - Q_{\theta_i}(s,a)\big)^2\big], 
\end{equation}

where $D$ is a replay memory buffer, used to store $(s,a,r,s')$ trajectories.
Equation (\ref{loss_dqn}) is the mean-squared error between the current approximation of the Q-function and some maximized target value  $r + \gamma \max_{a'}Q_{\theta_i^-}(s',a')$. The training of DQN has been shown to be more stable, and, consequently, DQN has been effective for many RL problems, including RL-CO problems. 
\end{itemize}

 \subsubsection{Policy-based methods}
In contrast to the value-based methods that aim to find the optimal state-action value function $Q^*(s,a)$ and act greedily with respect to it to obtain the optimal policy $\pi^*$, policy-based methods attempt to directly find the optimal policy, represented by some parametric function $\pi^*_{\theta}$, by optimizing \eqref{opt_pi} with respect to the policy parameters $\theta$: the method collects experiences in the environment using the current policy and optimizes it utilizing these collected experiences. Many methods have been proposed to optimize the policy functions, and we discuss the most commonly used ones for solving CO problems.

\begin{itemize}
    \item \textbf{Policy gradient.} In order to optimize (\ref{opt_pi}) with respect to the policy parameters $\theta$, policy gradient theorem \cite{sutton2000policy} can be applied to estimate the gradients of the policy function in the following form:
\begin{equation}
\label{eq:policygradient}
    \nabla_{\theta}J(\pi_{\theta})=\mathbb{E}_{\pi_{\theta}}[\sum\limits_{t=0}^{H} \nabla_{\theta}\log{\pi_{\theta}(a_t|s_t)}\hat{A}(s_t,a_t)],
\end{equation}
where \textit{the return estimate} $\hat{A}(s_t,a_t) = \sum\limits_{t=t'}^{H} \gamma^{t'-t} r(s_t', a_t') - b(s_t)$, $H$ is the agent's horizon, and $b(s)$ is the baseline function. The gradient of the policy is then used by the gradient descent algorithm to optimize the parameters $\theta$. 

\item \textbf{REINFORCE.} The role of the baseline $b(s)$ is to reduce the variance of the return estimate $\hat{A}(s_t,a_t)$ --- as it is computed by running the current policy $\pi_{\theta}$, the initial parameters can lead to poor performance in the beginning of the training, and the baseline $b(s)$ tries to mitigate this by reducing the variance. When the \textit{baseline} $b(s_t)$ is excluded from the return estimate calculation we obtain a \textit{REINFORCE} algorithm that has been proposed by \cite{Williams:92}. Alternatively, one can compute the baseline value $b(s_t)$ by calculating an average reward over the sampled trajectories, or by using a parametric value function estimator $V_{\phi}(s_t)$.

\item \textbf{Actor-critic algorithms.} The family of Actor-Critic (A2C, A3C) \cite{mnih2016asynchronous} algorithms further extend REINFORCE with the baseline by using bootstrapping --- updating the state-value estimates from the values of the subsequent states. For example, a common approach is to compute the return estimate for each step using the parametric value function:
\begin{equation}
    \hat{A}(s_t,a_t) = r(s_t, a_t) + V_{\phi}(s_t') - V_{\phi}(s_t)
\end{equation}

Although this approach introduces bias to the gradient estimates, it often reduces variance even further. Moreover, the actor-critic methods can be applied to the online and continual learning, as they no longer rely on Monte-Carlo rollouts, i.e. unrolling the trajectory to a terminal state.

\item \textbf{PPO/DDPG.} Further development of this group of reinforcement learning algorithms has resulted in the appearance of several more advanced methods such as \textit{Proximal Policy Optimization} (PPO) \cite{schulman2017proximal}, that performs policy updates with constraints in the policy space, or \textit{Deep Deterministic Policy Gradient} (DDPG) \cite{lillicrap2015continuous}, an actor-critic algorithm that attempts to learn a parametric state-action value function $Q_{\phi}(s,a)$, corresponding to the current policy, and use it to compute the bootstrapped return estimate. 


\end{itemize}

\subsubsection{Monte Carlo Tree Search}

\begin{figure}[h]
\centering
\includegraphics[width=\textwidth]{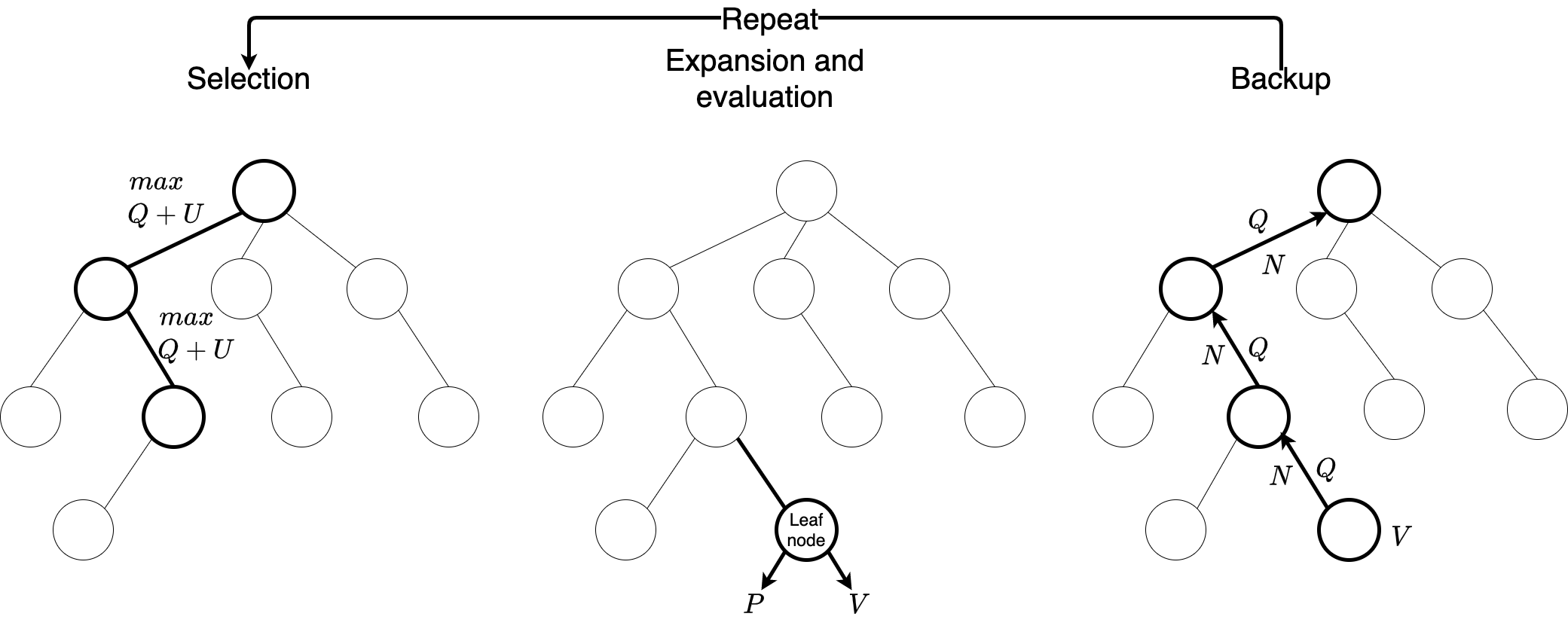}
\caption{Three steps of the Monte Carlo Tree Search (MCTS). Starting the simulation from the root node the \emph{select} step picks the node that maximizes the upper confidence bound. When previously unseen node is \emph{expanded}, the policy $P(s,*)$ and the state-value function $V(s)$ are \emph{evaluated} at this node, and the action-value $Q(s,a)$ and the counter $N(s,a)$ are initialized to $0$. Then $V(s)$ estimates are propagated back along the path of the current simulation to update $Q(s,a)$ and $N(s,a)$.}
\label{NMCTS}
\end{figure}

Both value-based and policy-based approaches do not use the model of the environment (model-free approaches), i.e. the transition probabilities of the model, and, hence, such approaches do not plan ahead by unrolling the environment to the next steps. However, it is possible to define an MDP for CO problems in such a way that we can use the knowledge of the environment in order to improve the predictions by planning several steps ahead. Some notable examples are AlphaZero \cite{silver2016mastering} and Expert Iteration \cite{10.5555/3295222.3295288} that have achieved super-human performances in games like chess, shogi, go, and hex, learning exclusively through self-play. Moreover, the most recent algorithm, MuZero \cite{schrittwieser2019mastering}, has been able to achieve a superhuman performance by extending the previous approaches using the learned dynamics model in challenging and visually complex domains, such as Atari games, go and shogi without the knowledge of the game rules.

\begin{itemize}
    \item \textbf{MCTS.} The algorithm follows the general procedure of Monte Carlo Tree Search (MCTS) \cite{mcts_survey} consisting of selection, expansion, roll-out and backup steps (Figure \ref{NMCTS}). However, instead of evaluating leaf nodes in a tree by making a rollout step, a neural network $f_{\theta}$ is used to provide a policy $P(s,*)$ and state-value estimates $V(s)$ for the new node in the tree. The nodes in the tree refer to states $s$, and edges refer to actions $a$. During the selection phase we start at a root state, $s_0$, and keep selecting next states that maximize the upper confidence bound:
    \begin{equation}
        \text{UCB} = Q(s,a) + c \cdot P(s,a) \cdot \frac{\sqrt{\sum_{a'} N(s,a')}}{1 + N(s,a)},
    \end{equation}
    When previously unseen node in a search tree is encountered, policy $P(s,*)$ and state-value value functions $P(s,*)$ and state-value estimates $V(s)$ are estimated for this node. After that $V(s)$ estimate is propagated back along the search tree updating the $Q(s,a)$ and $N(s,a)$ values. After a number of search iterations we select the next action from the root state according to the improved policy:
\begin{equation}
    \pi(s_o) = \frac{N(s_0,a)}{\sum_{a'} N(s_0, a')}.
\end{equation}
\end{itemize}

\section{Taxonomy of RL for CO}
\label{sec:taxonomy}

The full taxonomy of RL methods for CO can be challenging because of the orthogonality of the ways we can classify the given works. In this section we will list all the taxonomy groups that are used in this survey.

One straightforward way of dividing the RL approaches, concerning the CO field, is by the family of the RL methods used to find the solution of the given problem. As shown in the Figure~\ref{fig:rl_methods}, it is possible to split the RL methods by either the first level of the Figure~\ref{fig:rl_methods} (i.e. into the \textit{model-based} and \textit{model-free} methods) or by the second level (i.e. \textit{policy-based}, \textit{value-baded}, the methods using \textit{Monte-Carlo} approach) (section~\ref{sec:rl-algo}). In addition, another division is possible by the type of encoders used for representing the states of the MDP (section~\ref{subsec:encoders}). This division is much more granular than the other ones discussed in this section, as can be seen from the works surveyed in the next section.

Another way to aggregate the existing RL approaches is based on the integration of RL into the given CO problem, i.e. if an RL agent is searching for a solution to CO problem or if an RL agent is facilitating the inner workings of the existing off-the-shelf solvers. 
\begin{itemize}

\item In \textit{principal} learning an agent makes the direct decision that constitutes a part of the solution or the complete solution of the problem and does not require the feedback from the off-the-shelf solver. For example, in TSP the agent can be parameterized by a neural network that incrementally builds a path from a set of vertices and then receives the reward in the form of the length of the constructed path, which is used to update the policy of the agent. 

\item Alternatively one can learn the RL agent's policy in the \textit{joint} training with already existing solvers so that it can improve some of the metrics for a particular problem. For example, in MILP a commonly used approach is the Branch \& Bound method, which at every step selects a branching rule on the node of the tree. This can have a significant impact on the overall size of the tree and, hence, the running time of the algorithm. A branching rule is a heuristic that typically requires either some domain expertise or a hyperparameter tuning procedure. However, a parameterized RL agent can learn to imitate the policy of the node selection by receiving rewards proportional to the running time.
\end{itemize}

Another dimension that the RL approaches can be divided into is the way the solution is searched by the learned heuristics. In this regard, methods can be divided into those learning \emph{construction} heuristics or \emph{improvement} heuristics. 
\begin{itemize}
    \item Methods that learn \emph{construction} heuristics are building the solutions incrementally using the learned policy by choosing each element to add to a partial solution. 
    \item The second group of methods start from some arbitrary solution and learn a policy that \textit{improves} it iteratively. This approach tries to address the problem that is commonly encountered with the construction heuristics learning, namely, the need to use some extra procedures to find a good solution like beam search or sampling.  

\end{itemize}

\section{RL for CO}
\label{sec:problems}

In this section we survey existing RL approaches to solve CO problems that include Traveling Salesman Problem (Definition~\ref{def:tsp}), Maximum Cut Problem (Definition~\ref{def:maxcut}), Bin Packing Problem (Definition~\ref{def:3dbpp}), Minimum Vertex Cover Problem (Definition~\ref{def:minvertex}), and Maximum Independent Set (Definition~\ref{def:mis}). These problems have received the most attention from the research community and we juxtapose the approaches for all considered problems.

\subsection{Travelling Salesman Problem}

\begin{table*}[h]
\begin{center}
\centering
\footnotesize
\begin{tabular}{l|cc|cc}
\hline
& \multicolumn{2}{c|}{\textbf{\textit{Searching Solution}}} &
\multicolumn{2}{c}{\textbf{\textit{Training}}} \\
 Approach & Joint & Constructive & Encoder & RL \\
 \midrule
 \midrule
\cite{bello2016neural} & No & Yes & Pointer Network & REINFORCE with baseline \\
\cite{khalil2017learning} & No  & Yes & S2V & DQN\\
\cite{nazari2018reinforcement} & No  & Yes & Pointer Network with Convolutional Encoder & REINFORCE (TSP) and A3C (VRP) \\
\cite{inbook} & No &  Yes & Pointer Network with Attention Encoder & REINFORCE with baseline\\
\cite{kool2018attention} & No & Yes & Pointer Network with Attention Encoder & REINFORCE with baseline \\
\cite{emami2018learning} & No & No & FF NN with Sinkhorn layer & Sinkhorn Policy Gradient \\
\cite{cappart2020combining} & Yes & Yes & GAT/Set Transformer & DQN/PPO \\
\cite{drori2020learning} & Yes &  Yes & GIN with an Attention Decoder & MCTS \\
\cite{Lu2020A} & Yes & No  & GAT  & REINFORCE \\
\cite{chen2018learning}& Yes & No & LSTM encoder + classifier &  Q-Actor-Critic \\
\midrule
\end{tabular}
\end{center}
\caption{Summary of approaches for Travelling Salesman Problem.}
\label{tab:tsp}
\end{table*}

One of the first attempts to apply policy gradient algorithms to combinatorial optimization problems has been made in \cite{bello2016neural}. In the case of solving the Traveling Salesman Problem, the MDP representation takes the following form: a state is a $p$-dimensional graph embedding vector, representing the current tour of the nodes at the time step $t$, while the action is picking another node, which has not been used at the current state. This way the initial state $s_0$ is the embedding of the starting node. A transition function $T(s,a,s')$, in this case, returns the next node of the constructed tour until all the nodes have been visited. Finally, the reward in \cite{bello2016neural} is intuitive: it is the negative tour length. The pointer network architecture, proposed in \cite{vinyals2015pointer}, is used to encode the input sequence, while the solution is constructed sequentially from a distribution over the input using the pointer mechanism of the decoder, and trained in parallel and asynchronously similar to \cite{mnih2016asynchronous}. Moreover, several inference strategies are proposed to construct a solution --- along with greedy decoding and sampling Active Search approach is suggested. Active Search allows learning the solution for the single test problem instance, either starting from a trained or untrained model. To update the parameters of the controller so that to maximize the expected rewards REINFORCE algorithm with learned baseline is used.

The later works, such as the one by \cite{khalil2017learning}, has improved on the work of \cite{bello2016neural}.  In the case of \cite{khalil2017learning} the MDP, constructed for solving the Traveling Salesman problem, is similar to the one used by \cite{bello2016neural}, except for the reward function $r(s,a)$. The reward, in this case, is defined as the difference in the cost functions after transitioning from the state $s$ to the state $s'$ when taking some action $a$: $r(s, a) = c(h(s'), G)- c(h(s), G)$, where $h$ is the graph embedding function of the partial solutions $s$ and $s'$, $G$ is the whole graph, $c$ is the cost function. Because the weighted variant of the TSP is solved, the authors define a cost function $c(h(s), G)$ as the negative weighted sum of the tour length. Also, the work implements S2V \cite{dai2016discriminative} for encoding the partial solutions, and DQN as the RL algorithm of choice for updating the network's parameters.

Another more work by \cite{nazari2018reinforcement}, motivated by \cite{bello2016neural}, concentrates on solving the Vehicle Routing Problem (VRP), which is a generalization of TSP. However, the approach suggested in \cite{bello2016neural} can not be applied directly to solve VRP due to its dynamic nature, i.e. the demand in the node becoming zero, once the node has been visited since it embeds the sequential and static nature of the input. The authors of \cite{nazari2018reinforcement} extend the previous methods used for solving TSP to circumvent this problem and find the solutions to VRP and its stochastic variant. Specifically, similarly to \cite{bello2016neural}, in \cite{nazari2018reinforcement} approach the state $s$ represents the embedding of the current solution as a vector of tuples, one value of which is the coordinates of the customer's location and the other is the customer's demand at the current time step. An action $a$ is picking a node, which the vehicle will visit next in its route. The reward is also similar to the one used for TSP: it is the negative total route length, which is given to the agent only after all customers' demands are satisfied, which is the terminal state of the MDP. The authors of \cite{nazari2018reinforcement} also suggest to improve the Pointer Network, used by \cite{bello2016neural}. To do that, the encoder is simplified by replacing the LSTM unit with the 1-d convolutional embedding layers so that the model is invariant to the input sequence order, consequently, being able to handle the dynamic state change. The policy learning is then performed by using REINFORCE algorithm for TSP and VRP while using A3C for stochastic VRP.

Similarly to \cite{nazari2018reinforcement}, the work by \cite{inbook} uses the same approach as the one by \cite{bello2016neural}, while changing the encoder-decoder network architecture. This way, while the MDP is the same as in \cite{bello2016neural}, instead of including the LSTM units, the GNN encoder architecture is based solely on the attention mechanisms so that the input is encoded as a set and not as a sequence. The decoder, however, stays the same as in the Pointer Network case. Additionally, the authors have looked into combining a solution provided by the reinforcement learning agent with the 2-Opt heuristic \cite{Croes58}, in order to further improve the inference results. REINFORCE algorithm with critic baseline is used to update the parameters of the described encode-decoder network. 

Parallel to \cite{inbook}, inspired by the transformer architecture of \cite{vaswani2017attention}, a \textit{construction} heuristic learning approach by \cite{kool2018attention} has been proposed in order to solve TSP, two variants of VRP (Capacitated VRP and Split Delivery VRP), Orienteering Problem (OP), Prize Collecting TSP (PCTSP) and Stochastic PCTSP (SPCTSP). 
In this work, the authors have implemented a similar encoder-decoder architecture as the authors of \cite{inbook}, i.e. the Transformer-like attention-based encoder, while the decoder is similar to one of the Pointer Network. However, the authors found that slightly changing the training procedure and using a simple rollout baseline instead of the one learned by a critic yields better performance. The MDP formulation, in this case, is also similar to the one used by \cite{inbook}, and, consequently, the one by \cite{bello2016neural}.

One specific \textit{construction} heuristic approach has been proposed in \cite{emami2018learning}. The authors have designed a novel policy gradient method, Sinkhorn Policy Gradient (SPG), specifically for the class of combinatorial optimization problems, which involves permutations. This approach yields a different MDP formulation. Here, in contrast with the case when the solution is constructed sequentially, the state space consists of instances of combinatorial problems of a particular size. The action space, in this case, is outputting a permutation matrix, which, applied to the original graph, produces the solution tour. The reward function is the negated sum of the Euclidean distances between each stop along the tour. Finally, using a special Sinkhorn layer on the output of the feed-forward neural network with GRUs to produce continuous and differentiable relaxations of permutation matrices, authors have been able to train actor-critic algorithms similar to Deep Deterministic Policy Gradient (DDPG) \cite{lillicrap2015continuous}.

The work by \cite{cappart2020combining} combines two approaches to solving the traveling salesman problem with time windows, namely the RL approach and the constraint programming (CP) one, so that to learn branching strategies. In order to encode the CO problems, the authors bring up a dynamic programming formulation, that acts as a bridge between both techniques and can be exposed both as an MDP and a CP problem. A state $s$ is a vector, consisting of three values: the set of remaining cities that still have to be visited, the last city that has been visited, and the current time. An action $a$ corresponds to choosing a city. The reward $r(s,a)$ corresponds to the negative travel time between two cities. This MDP can then be transformed into a dynamic programming model. DQN and PPO algorithms have been trained for the MDP formulation to select the efficient branching policies for different CP search strategies --- branch-and-bound, iterative limited discrepancy search and restart based search, and have been used to solve challenging CO problems.

The work by \cite{drori2020learning} differs from the previous works, which tailor their approaches to individual problems. In contrast, this work provides a general framework for model-free reinforcement learning using a GNN representation that adapts to different problem classes by changing a reward. This framework models problems by using the edge-to-vertex line graph and formulates them as a single-player game framework. The MDPs for TSP and VRP are the same as in \cite{bello2016neural}. Instead of using a full-featured Neural MCTS, \cite{drori2020learning} represents a policy as a GIN encoder with an attention-based decoder, learning it during the tree-search procedure.

\cite{Lu2020A} suggests to learn the \textit{improvement} heuristics in hierarchical manner for capacitated VRP as a part of the \textit{joint} approach. The authors have designed an intrinsic MDP, which incorporates not only the features of the current solutions but also the running history. A state $s_i$ includes free capacity of the route containing a customer $i$, its location, a location of the node $i^-$ visited before $i$, a location of the node $i^+$ visited after $i$,a distance from $i^-$ to $i$,a distance from $i$ to $i^+$,a distance from $i^-$ to $i^+$, an action taken $h$ steps before, an effect of $a_{t-h}$. The action consists of choosing between two groups of operators, that change the current solution, for example by applying the 2-Opt heuristic, which removes two edges and reconnects their endpoints. Concretely, these two operator groups are improvement operators, that are chosen according to a learned policy, or perturbation operators in the case of reaching a local minima. The authors have experimented with the reward functions, and have chosen the two most successful ones: $+1/-1$ reward for each time the solution improves/does not give any gains, and the advantage reward, which takes the initial solution's total distance as the baseline, and constitutes the difference between this baseline and the distance of the subsequent solutions as the reward at each time step. The policy is parameterized by the Graph Attention Network and is trained with the REINFORCE algorithm.

The final work, we are going to cover for this section of problems is by \cite{chen2018learning}, who proposes solving VRP and online job scheduling problems by learning \textit{improvement} heuristics. The algorithm rewrites different parts of the solution until convergence instead of constructing the solution in the sequential order. The state space is represented as a set of all solutions to the problem, while the action set consists of regions, i.e. nodes in the graph, and their corresponding rewriting rules. The reward, in this case, is the difference in the costs of the current and previous solutions. The authors use an LSTM encoder, specific to each of the covered problems and train region-picking and rule-picking policies jointly by applying the Q-Actor-Critic algorithm.

\subsection{Maximum Cut Problem} \label{MaxCut}

\begin{table*}[h]
\begin{center}
\centering
\footnotesize
\begin{tabular}{l|cc|cc}
\hline
& \multicolumn{2}{c|}{\textbf{\textit{Searching Solution}}} &
\multicolumn{2}{c}{\textbf{\textit{Training}}} \\
 Approach & Joint & Constructive & Encoder & RL \\
 \midrule
 \midrule
\cite{khalil2017learning} & No & Yes & S2V & DQN \\
\cite{barrett2019exploratory} & No & Yes & S2V & DQN \\
\cite{cappart2019improving} & Yes & Yes & S2V & DQN \\
\cite{tang2019reinforcement} & Yes & No & LSTM + Attention & Policy Gradient + ES \\
 \cite{abe2019solving} & No & Yes & GNN & Neural MCTS \\
\cite{Gu_2020} & No & Yes & Pointer Network & A3C \\

\midrule
\end{tabular}
\end{center}
\caption{Summary of approaches for Maximum Cut Problem.}\label{tab:maxcut}
\end{table*}

The first work to address solving Maximum Cut Problem with reinforcement learning was \cite{khalil2017learning}, that proposed the \emph{principled} approach to learning the \emph{construction} heuristic by combining graph embeddings with Q-learning - S2V-DQN. They formulated the problem as an MDP, where the state space, $S$, is defined as a partial solution to the problem, i.e. the subset of all nodes in a graph added to the set, that maximizes the maximum cut. The action space, $A$, is a set of nodes that are not in the current state. The transition function, $T({s}_{t+1} | {s}_t, {a}_t)$, is deterministic and corresponds to tagging the last selected node with a feature $x_v = 1$. The reward is calculated as an immediate change in the cut weight, and the episode terminates when the cut weight can't be improved with further actions. The graph embedding network proposed in \cite{dai2016discriminative} was used as state encoder. A variant of the Q-learning algorithm was used to learn to construct the solution, that was trained on randomly generated instances of graphs. This approach achieves better approximation ratios compared to the commonly used heuristic solutions of the problem, as well as the generalization ability, which has been shown by training on graphs of consisting of 50-100 nodes and tested on graphs with up to 1000-1200 nodes, achieving very good approximation ratios to exact solutions.

\cite{barrett2019exploratory} improved on the work of \cite{khalil2017learning} in terms of the approximation ratio as well as the generalization by proposing an ECO-DQN algorithm. The algorithm kept the general framework of S2V-DQN but introduced several modifications. The agent was allowed to remove vertices from the partially constructed solution to better explore the solution space. The reward function was modified to provide a normalized incremental reward for finding a solution better than seen in the episode so far, as well as give small rewards for finding a locally optimal solution that had not yet been seen during the episode. In addition, there were no penalties for decreasing cut value. The input of the state encoder was modified to account for changes in the reward structure. Since in this setting the agent had been able to  explore indefinitely, the episode length was set to $2|V|$. Moreover, the authors allowed the algorithm to start from an arbitrary state, which could be useful by combining this approach with other methods, e.g. heuristics. This method showed better approximation ratios than S2V-DQN, as well as better generalization ability.

\cite{cappart2019improving} devised the \emph{joint} approach to the Max-Cut problem by incorporation of the reinforcement learning into the Decision Diagrams (DD) framework \cite{10.5555/3028836} to learn the \emph{constructive} heuristic. The integration of the reinforcement learning allowed to provide tighter objective function bounds of the DD solution by learning heuristics for variable ordering. They have formulated the problem as an MDP, where the state space, $S$, is represented as a set of ordered sequences of selected variables along with partially constructed DDs. The action space, $A$, consists of variables, that are not yet selected. The transition function, $T$, adds variables to the selected variables set and to the DD. The reward function is designed to tighten the bounds of the DD and is encoded as the relative upper and lower bounds improvements after the addition of the variable to the set. The training was performed on the generated random graphs with the algorithm and state encoding described above in \cite{khalil2017learning}. The authors showed that their approach had outperformed several ordering heuristics and generalized well to the larger graph instances, but didn't report any comparison to the other reinforcement learning-based methods.

Another \emph{joint} method proposed by \cite{tang2019reinforcement} combined a reinforcement learning framework with the cutting plane method. Specifically, in order to learn the \emph{improvement} heuristics to choose Gomory’s cutting plane, which is frequently used in the Branch-and-Cut solvers, an efficient MDP formulation was developed. The state space, $S$, includes the original linear constraints and cuts added so far. Solving the linear relaxation produces the action space, $A$, of Gomory cut that can be added to the problem. After that the transition function, $T$, adds the chosen cuts to the problem that results in a new state. The reward function is defined as a difference between the objective function of two consecutive linear problem solutions. The policy gradient algorithm was used to select new Gamory cuts, and the state was encoded with an LSTM network (to account for a variable number of variables) along with the attention-based mechanism (to account for a variable number of constraints). The algorithm was trained on the generated graph instances using evolution strategies and had been shown to improve the efficiency of the cuts, the integrality gaps, and the generalization, compared to the usually used heuristics used to choose Gomory cuts. Also, the approach was shown to be beneficial in combination with the branching strategy innexperiments with the Branch-and-Cut algorithm.

\cite{abe2019solving} proposed to use a graph neural network along with Neural MTCS approach to learn the \emph{construction} heuristics. The MDP formulation defines the state space, $S$, as a set of partial graphs from where nodes can be removed and colored in one of two colors representing two subsets. The action space, $A$, represents sets of nodes still left in the graph and their available colors. The transition function, $T$, colors the selected node of the graph and removes it along with the adjacent edges. The neighboring nodes, that have been left, are keeping a counter of how many nodes of the adjacent color have been removed. When the new node is removed, the number of the earlier removed neighboring nodes of the opposite color is provided as the incremental reward signal, $R$ (the number of edges that were included in the cut set). Several GNNs were compared as the graph encoders, with GIN\cite{xu2018powerful} being shown to be the most performing. Also, the training procedure similar to AlphaGo Zero was employed with the modification to accommodate for a numeric rather than win/lose solution. The experiments were performed with a vast variety of generated and real-world graphs. The extensive comparison of the method with several heuristics and with previously described S2V-DQN \cite{khalil2017learning} showed the superior performance as well as the better generalization ability to larger graphs, yet they didn't report any comparison with the exact methods.

\cite{Gu_2020} applied the Pointer Network \cite{vinyals2015pointer} along with the Actor-Critic algorithm similar to \cite{bello2016neural}\ to \emph{iteratively} construct a solution. The MDP formulation defines the state, $S$, as a symmetric matrix, $Q$, the values of which are the edge weights between nodes (0 for the disconnected nodes). Columns of this matrix are fed to the Pointer Network, which sequentially outputs the actions, $A$, in the form of pointers to input vectors along with a special end-of-sequence symbol "EOS". The resulting sequence of nodes separated by the "EOS" symbol represents a solution to the problem, from which the reward is calculated. The authors conducted experiments with simulated graphs with up to 300 nodes and reported fairly good approximations ratios, but, unfortunately, didn't compare with the previous works or known heuristics.

\subsection{Bin Packing Problem}

\begin{table*}[h]
\begin{center}
\centering
\footnotesize
\begin{tabular}{l|cc|cc}
\hline
& \multicolumn{2}{c|}{\textbf{\textit{Searching Solution}}} &
\multicolumn{2}{c}{\textbf{\textit{Training}}} \\
 Approach & Joint & Constructive & Encoder & RL \\
 \midrule
 \midrule
\cite{hu2017solving} & No & Yes & Pointer Network & REINFORCE with baseline \\
\cite{duan2018multitask} & No & Yes & Pointer Network + Classifier & PPO \\
\cite{alex2018ranked} & No & Yes & FF NN & Neural MCTS \\
\cite{li2020solving} & No & No & Attention & Actor-Critic \\
\cite{cai2019reinforcement} & Yes & No & N/A & PPO \\

\midrule
\end{tabular}
\end{center}
\caption{Summary of approaches for Bin Packing Problem.}\label{tab:bin}
\end{table*}

To our knowledge, one of the first attempts to solve a variant of Bin Packing Problem with modern reinforcement learning was \cite{hu2017solving}. The authors have proposed a new, more realistic formulation of the problem, where the bin with the least surface area that could pack all 3D items is determined. This \emph{principled} approach is only concerned with learning the \emph{construction} heuristic to choose a better sequence to pack the items and using regular heuristics to determine the space and orientation. The state space, $S$, is denoted by a set of sizes (height, width, and length) of the items that need to be packed. The approach proposed by \cite{bello2016neural}, which utilizes the Pointer Network, is used to output the sequence of actions, $A$, i.e. sequence of items to pack. Reward, $R$, is calculated as the value of the surface area of packed items. REINFORCE with the baseline is used as a reinforcement learning algorithm, with the baseline provided by the known heuristic. The improvement over the heuristic and random item selection was shown with greedy decoding as well as sampling from the with beam search.

Further work by \cite{duan2018multitask} extends the approach of \cite{hu2017solving} to learning of the orientations along with a sequence order of items by combining reinforcement and supervised learning in a multi-task fashion. In this work a Pointer Network, trained with a PPO algorithm, was enhanced with a classifier that determined the orientation of the current item in the output sequence, given the representation from the encoder and the embedded partial items sequence. The classifier is trained in a supervised setting, using the orientations in the so-far best solution of the problem as labels. The experiments were conducted on the real-world dataset and showed that the proposed method performs better than several widely used heuristics and previous approaches by \cite{hu2017solving}.

\cite{alex2018ranked} applied the \emph{principled} Neural MCTS approach to solve the already mentioned variant of 2D and 3D bin packing problems by learning the \emph{construction} heuristic. The MDP formulation includes the state space, $S$, represented by the set of items that need to be packed with their heights, widths, and depths. The action space, $A$, is represented by the set of item ids, coordinates of the bottom-left corner of the position of the items, and their orientations. To solve 2D and 3D Bin Packing Problems, formulated as a single-player game, Neural MCTS constructs the optimal solution with the addition of a ranked reward mechanism that reshapes the rewards according to the relative performance in the recent games. This mechanism aims to provide a natural curriculum for a single agent similar to the natural adversary in two-player games. The experimental results have been compared with the heuristic as well as Gurobi solver and showed better performance in several cases on the dataset created by randomly cutting the original bin into items. 

\cite{li2020solving} tries to address the limitation of the three previously described works, namely using heuristics for the rotation or the position coordinates (\cite{hu2017solving},\cite{duan2018multitask}) or obtaining items from cutting the original bin (\cite{alex2018ranked}). Concretely, the authors propose to construct an end-to-end pipeline to choose an item, orientation, and position coordinates by using an attention mechanism. The MDP's state space, $S$, includes a binary indicator of whether the item is packed or not, its dimensions, and coordinates relative to the bin. The action space, $A$, is defined by the selection of the item, the rotation and the position of the item in the bin. The reward function is incremental and is calculated as the volume gap in the bin, i.e. the current bin's volume $-$ the volume of the packed items. The actor-critic algorithm was used for learning. The comparison provided with a genetic algorithm and previous reinforcement learning approaches, namely \cite{duan2018multitask} and \cite{kool2018attention}, has showed that the proposed method achieves a smaller bin gap ratio for the problems of size up to 30 items.

\cite{cai2019reinforcement} has taken the \emph{joint} approach to solving a 1D bin packing problem by combining proximal policy optimization (PPO) with the simulated annealing (SA) heuristic algorithm. PPO is used to learn the \emph{improvement} heuristic to build an initial starting solution for SA, which in turn, after finding a good solution in a limited number of iterations, calculates the reward function, $R$, as the difference in costs between the initial and final solutions and passes it to the PPO agent. The action space, $A$, is represented by a set of changes of the bins between two items, e.g. a perturbation to the current solution. The state space, $S$, is described with a set of assignments of items to bins. The work has showed that the combination of RL and the heuristics can find solutions better than these algorithms in isolation, but has not provides any comparison with known heuristics or other algorithms.

\subsection{Minimum Vertex Cover Problem}
\label{sec:mvc}
\begin{table*}[h]
\begin{center}
\centering
\footnotesize
\begin{tabular}{l|cc|cc}
\hline
& \multicolumn{2}{c|}{\textbf{\textit{Searching Solution}}} &
\multicolumn{2}{c}{\textbf{\textit{Training}}} \\
 Approach & Joint & Constructive & Encoder & RL \\
 \midrule
 \midrule
\cite{khalil2017learning} & No & Yes & S2V & DQN \\
\cite{song2019co} & No & Yes & S2V & DQN + Imitation Learning \\
\cite{DBLP:journals/corr/abs-1903-03332} & No & Yes & GNN & DQN \\

\midrule
\end{tabular}
\end{center}
\caption{Summary of approaches for Minimum Vertex Cover problem.}\label{tab:mvc}
\end{table*}

The \emph{principled} approach to solving the Minimum Vertex Problem (MVC) with reinforcement learning was developed by \cite{khalil2017learning}. To learn the \emph{construction} heuristic the problem was put into the MDP framework, which is described in details in \ref{MaxCut} along with the experimental results. To apply the algorithm to the MVC problem reward function, $R$, was modified to produce $-1$ for assigning a node to the cover set. Episode termination happens when all edges are covered.  

\cite{song2019co} proposed the \emph{joint} co-training method, which has gained popularity in the classification domain, to construct sequential policies. The paper describes two policy-learning strategies for the MVC problem: the first strategy copies the one described in \cite{khalil2017learning}, i.e. S2V-DQN, the second is the integer linear programming approach solved by the branch \& bound method. The authors create the $CoPiEr$ algorithm that is intuitively similar to Imitation Learning \cite{hester2018deep}, in which two strategies induce two policies, estimate them to figure out which one is better, exchange the information between them, and, finally, make the update. The performed experiments resulted in the extensive ablation study, listing the comparisons with the S2V-DQN, Imitation learning, and Gurobi solver, and showed a smaller optimality gap for problems up to 500 nodes.

Finally, it is worth including in this section the work by \cite{DBLP:journals/corr/abs-1903-03332} that combined the supervised and reinforcement learning approaches in the \emph{joint} method that learns a \emph{construction} heuristic for a budget-constrained Maximum Vertex Cover problem. The algorithm consists of two phases. In the first phase, a GCN is used to determine "good" candidate nodes by learning the scoring function, using the scores, provided by the probabilistic greedy approach, as labels. Then the candidates nodes are used in an algorithm similar to \cite{khalil2017learning} to sequentially construct a solution. Since the degree of nodes in large graphs can be very high, the importance sampling according to the computed score is used to choose the neighboring nodes for the embedding calculation, which helps to reduce the computational complexity. The extensive experiments on random and real-world graphs has showed that the proposed method performs marginally better compared to S2V-DQN, scales to much larger graph instances up to a hundred thousand nodes, and is significantly more efficient in terms of the computation efficiency due to a lower number of learned parameters.

\subsection{Maximum Independent Set Problem}
One of the first RL for CO works, that covered MIS problem, is \cite{cappart2019improving}. It focuses on a particular approach to solving combinatorial optimization problems, where an RL algorithm is used to find the optimal ordering of the variables in a Decision Diagram (DD), to tighten the relaxation bounds for the MIS problem. The MDP formulation, as well as the encoder and the RL algorithm are described in detail in Section \ref{sec:mvc}.

Another early article, covering MIS, is \cite{abe2019solving}. In it, authors have proposed the following MDP formulation: let a state $s \in S$ be a graph, received at each step of constructing a solution, with the initial state $s_0$ being the initial graph $G_0$; an action $a \in A$ be a selection of one node of a graph in the current state; a transition function $T(s,a,s')$ be the function returning the next state, corresponding to the graph where edges covered by the action $a$ and its adjacent nodes are deleted; and, finally, a reward function $r(s,a)$ be a constant and equal to $1$. For encoders, the authors proposed to apply a GIN, to account for the variable size of a state representation in a search tree. \cite{abe2019solving} uses a model-based algorithm, namely AlphaGo Zero, to update the parameters of the GIN network.

\begin{table*}[h]
\begin{center}
\centering
\footnotesize
\begin{tabular}{l|cc|cc}
\hline
& \multicolumn{2}{c|}{\textbf{\textit{Searching Solution}}} &
\multicolumn{2}{c}{\textbf{\textit{Training}}} \\
 Approach & Joint & Constructive & Encoder & RL \\
 \midrule
 \midrule
\cite{cappart2019improving} & Yes & No & S2V & DQN \\
\cite{abe2019solving}& No & No & GIN &MCTS \\
\cite{ahn2020deep} & Yes & Yes & GCN & PPO \\
\midrule
\end{tabular}
\end{center}
\caption{Summary of approaches for Maximum Independent Set problem.}\label{tab:mis}
\end{table*}

The latest work, \cite{ahn2020deep} modifies the MDP formulation, by applying a label to each node, i.e. each node can be either included into the solution, excluded from the solution or the determination of its label can be deferred. This way, a state $s \in S$  becomes a vector, the size of which is equal to the number of nodes in the graph, and which consists of the labels that have been given to each node at the current time step. The initial state $s_0$ is a vector with all labels set to being deferred. An action  $a \in A$ is a vector with new label assignments for the next state of only currently deferred nodes. To maintain the independence of the solution set, the transition function $T(s,a)$ is set to consist of two phases: the update phase and clean-up phase. The first phase represents the naive assignment of the labels by applying the action $a$, which leads to the intermediate state $\hat{s}$. In the clean-up phase, the authors modify the intermediate state $\hat{s}$ in such a way that the included nodes are only adjacent to the excluded ones. Finally, the reward function $r(s,a)$ is equal to the increase in the cardinality of included vertices between the current state $s'$ and the previous state $s$. The authors propose to use the Graph Convolutional Network encoder and PPO method with the rollback procedure to learn the optimal Deep Auto-Deferring Policy (ADP), which outputs the \textit{improvement} heuristic to solve the MIS problem.

\section{Comparison}
\label{sec:comparison}
\begin{table*}[t!]
\setlength\tabcolsep{4.5pt}
\renewcommand{\arraystretch}{1.5}
\centering
\resizebox{\textwidth}{!}{
\begin{tabular}{ |c|c|c|c|c|c| }

 \hline
 \multirow{2}{*}{Algo}&\multirow{2}{*}{Article}& \multirow{2}{*}{Method}&\multicolumn{3}{|c|}{Average tour length} \\
 \cline{4-6}
 &&&N=20& N=50 & N=100\\
 \cline{4-6}
 \hline
 
 \multirow{6}{*}{RL} &\cite{Lu2020A}& \multirow{3}{*}{REINFORCE}& 4.0   & 6.0& 8.4\\ 
&\cite{kool2018attention}&& 3.8 & 5.7 & 7.9\\

&\cite{inbook}&& 3.8 & 5.8 & 8.9\\	
\cline{3-6}
&\cite{inbook}& REINFORCE+2opt & 3.8 & 5.8	& 8.2\\
\cline{3-6}
&\cite{bello2016neural}& A3C& 3.8 & 5.7	& 7.9\\
\cline{3-6}
&\cite{emami2018learning}&Sinkhorn Policy Gradient& 4.6 & -- & -- \\
\midrule \midrule
&\cite{helsgaun2017extension}&LK-H& 3.8 & 5.7 & 7.8\\ 
&\cite{ortools}&OR-Tools& 3.9 & 5.8 & 8.0\\ 
 \hline
 \end{tabular}
 }
\caption{The average tour lengths (the smaller, the better) comparison for TSP for ER graphs with the number of nodes $N$ equal to 20, 50, 100.}
\label{tab:comp1}
\end{table*}

In this section, we will partially compare the results achieved by the works presented in this survey. Concretely, we have distinguished the two most frequently mentioned problems, namely, Travelling Salesman Problem (TSP) and Capacitated Vehicle Routing Problem (CVRP). The average tour lengths for these problems, reported in the works \cite{Lu2020A,kool2018attention,lodi2002heuristic,bello2016neural,emami2018learning,ma2019combinatorial,nazari2018reinforcement,chen2018learning}, are shown in Tables \ref{tab:comp1}, \ref{tab:comp2}. 

 The presented results have been achieved on Erdős–Rényi (ER) graphs of various sizes, namely, with the number of nodes of 20, 50, 100 for TSP, and 10, 20, 50, 100 for CVRP. In the case of CVRP, we have also specified the capacity of the vehicle (\text{Cap.}), which varies from 10 to 50. Also, we have included the results achieved by the OR-Tools solver \cite{ortools} and LK-H heuristic algorithm \cite{helsgaun2017extension} as the baseline solutions. 

\paragraph{Best performing methods.} It is clear from the presented table that the best performing methods for TSP are \cite{kool2018attention} and \cite{bello2016neural}, and for VRP --- \cite{Lu2020A}. These algorithms perform on par with the baseline, and in some cases demonstrate better results. Moreover, in the case of \cite{Lu2020A}, the algorithm manages to present the best performance across all the other methods, even for tasks with smaller vehicle capacities. 

\paragraph{Focus on smaller graphs.} Throughout our analysis, we have found that most of the articles focus on testing the CO-RL methods on graphs with the number of nodes of 20, 50, 100. At the same time, \cite{ma2019combinatorial} presents the results for bigger graphs with 250, 500, 750, and 1000 nodes for a TSP problem. This may be connected to the fact that with the increasing size of the graphs the process of finding the optimal solution also becomes much more computationally difficult even for the commercial solvers. The comparison of the reported results and the baseline further supports this fact: for TSP it can be seen how for smaller graphs almost all of the methods outperform OR-Tools, while for bigger graphs it is no longer the case. Consequently, this can be a promising direction for further research.

\paragraph{Non-overlapping problems.} We can see that although there have emerged a lot of works focused on creating well-performing RL-based solvers, the CO problems, covered in these articles, rarely coincide, which makes the fair comparison a much harder task. We are convinced that further analysis should be focused on unifying the results from different sources, and, hence, identifying more promising directions for research.

\paragraph{Running times.} One of the main pros of using machine learning and reinforcement learning algorithms to solve CO problems is the considerable reduction in running times compared to the ones obtained by the metaheuristic algorithms and solvers. However, it is still hard to compare the running time results of different works as they can significantly vary depending on the implementations and the hardware used for experimentation. For these reasons, we do not attempt to exactly compare the times achieved by different RL-CO works. Still, however, we can note that some of the works such as \cite{nazari2018reinforcement}, \cite{chen2018learning}, \cite{Lu2020A} claim to have outperformed the classic heuristic algorithms. Concretely, the authors of \cite{nazari2018reinforcement}, show that for larger problems, their framework is faster than the randomized heuristics and their running times grow slower with the increase of the complexities of the CO problems than the ones achieved by Clarke-Wright \cite{Clarke1964} and Sweep heuristics \cite{Wren1972}. \cite{chen2018learning} claim that their approach outperforms the expression simplification component in Z3 solver \cite{de2008z3} in terms of both the objective metrics and the time efficiency. Finally, although the exact training times are not given in the article, the authors of \cite{Lu2020A} note that the given time of their algorithm is much smaller than that of LK-H. In addition, although also acknowledging the complexity of comparing the times of different works, \cite{kool2018attention} have claimed that the running time of their algorithm is ten times faster than the one of \cite{bello2016neural}.

\begin{table*}[t!]
\setlength\tabcolsep{1.0pt}
\renewcommand{\arraystretch}{1.5}
\centering
\resizebox{\textwidth}{!}{
\begin{tabular}{ |c|c|c|c|c|c|c|c|c|c| }
  \hline
 \multirow{2}{*}{Algo}&\multirow{2}{*}{Article}& \multirow{2}{*}{Method}&\multicolumn{7}{|c|}{Average tour length} \\
 \cline{4-10}
 &&&N=10,&N=20,& N=20,&N=50,& N=50,& N=100,& N=100,\\
  &&&Cap. 10&Cap. 20 &Cap. 30 &Cap. 30 &Cap. 40 &Cap. 40 &Cap. 50\\
 \cline{4-10}
  \hline
 \hline
 \multirow{4}{*}{RL} &\cite{nazari2018reinforcement}&\multirow{3}{6em}{REINFORCE}
& 4.7 &--&	6.4 &--&	11.15 &--&	17.0\\
& \cite{kool2018attention}&
&--&--&	6.3 &--&	10.6 &--& 16.2 \\
 &\cite{Lu2020A}&
&--& 6.1	&--&	10.4&--&	15.6 &-- \\
 \cline{3-10}
 &\cite{chen2018learning}&A2C&--&--&	6.2 &--&	10.5 &--& 16.1 \\
\midrule \midrule
&\cite{helsgaun2017extension}&LK-H& -- & 6.1 & 6.1 &10.4 & 10.4 & 15.6 & 15.6\\ 
&\cite{ortools}&OR-Tools& 4.7 & 6.4 & 6.4 &11.3 & 11.3 & 17.2 & 17.2\\ 
 \hline
\end{tabular}
}
\caption{The average tour lengths comparison for Capacitated VRP for ER graphs with the number of nodes $N$ equal to 10, 20, 50, 100. \textit{Cap.} represents the capacity of the vehicle for CVRP.}
\label{tab:comp2}
\end{table*}

\section{Conclusion and future directions}
\label{sec:conclusion}

The previous sections have covered several approaches to solving canonical combinatorial optimization problems by utilizing reinforcement learning algorithms. As this field has demonstrated to be performing on-par with the state-of-the-art heuristic methods and solvers, we are expecting new algorithms and approaches to emerge in the following possible directions, which we have found promising:

 \paragraph{Generalization to other problems.} In \ref{sec:comparison}, we have formulated one of the main problems of the current state of the RL-CO field, which is a limited number of experimental comparisons. Indeed, the CO group of mathematical problems is vast, and the current approaches often require being implemented for a concrete set of problems. RL field, however, has already made some steps towards the generalization of the learned policies to the unseen problems (for example, \cite{GroshevGeneralization}). In the case of CO, these unseen problems can be smaller instances of the same problem, problem instances with different distributions, or even the ones from the other group of CO problems. We believe, that although this direction is challenging, it is extremely promising for future development in the RL-CO field.
 
\paragraph{Improving the solution quality.} A lot of the reviewed works, presented in this survey, have demonstrated superior performance compared to the commercial solvers. Moreover, some of them have also achieved the quality of the solutions equal to the optimal ones or the ones achieved by the heuristic algorithms. However, these results are true only for the less complex versions of CO problems, for example, the ones with smaller numbers of nodes. This leaves us with the possibility of further improvement of the current algorithms in terms of the objective quality. Some of the possible ways for this may be further incorporation of classical CO algorithms with the RL approaches, for example, with using imitation learning as in \cite{hester2018deep}.
\paragraph{Filling the gaps.} One of the ways to classify RL-CO approaches, which we have mentioned previously, is by grouping them into \textit{joint} and \textit{constructive} methods. Tables \ref{tab:tsp}, \ref{tab:maxcut}, \ref{tab:bin}, \ref{tab:mvc}, \ref{tab:mis} contain the information about these labels for each of the reviewed article, and from them, we can identify some unexplored approaches for each of the CO problems. This way from Table \ref{tab:bin}, it can be seen that there has not been published any both joint and constructive algorithm for solving the Bin Packing problem. The same logic can be applied to the Minimum Vertex Problem, Table \ref{tab:bin}, where there are no approaches of joint-constructive and joint-nonconstructive type. Exploring these algorithmic possibilities can provide us not only with the new methods but also with useful insights into the effectiveness of these approaches.

\bigskip

In conclusion, we see the field of RL for CO problems as a very promising direction for CO research, because of the effectiveness in terms of the solution quality, the capacity to outperform the existing algorithms, and huge running time gains compared to the classical heuristic approaches.






\bibliography{mybibfile}
\end{document}